\documentclass{article} % For LaTeX2e
\usepackage{iclr2027_conference,times}

\usepackage[utf8]{inputenc} % allow utf-8 input
\usepackage[T1]{fontenc}    % use 8-bit T1 fonts
\usepackage{hyperref}
\usepackage{url}
\usepackage{booktabs}
\usepackage{graphicx}
\usepackage{wrapfig}
\usepackage{fancyvrb}

\usepackage{microtype}      % microtypography
\usepackage{xcolor}         % colors
\usepackage[english]{babel}
\usepackage{enumitem}

\makeatletter
\renewcommand\paragraph{\@startsection{paragraph}{4}{\z@}%
  {1ex plus 0.5ex minus .2ex}%   keep ICLR's space before
  {-0.4em}%                     ~one interword space after
  {\normalsize\bf}}
\makeatother

\definecolor{hypercolor}{HTML}{0ea175}

\hypersetup{
    colorlinks=true,
    linkcolor=hypercolor,
    urlcolor=hypercolor,
    citecolor=hypercolor,
    filecolor=hypercolor,
}
\setlist{leftmargin=0.25in}

\usepackage{caption}
\usepackage{listings}
\usepackage{tcolorbox}
\usepackage{capt-of}
\usepackage{textcomp}
\tcbuselibrary{skins}
\definecolor{codestr}{RGB}{163,21,21}
\definecolor{codebg}{RGB}{245,247,250}
\definecolor{codeborder}{RGB}{210,215,222}
\definecolor{codekw}{RGB}{0,92,170}
\definecolor{codecomment}{RGB}{110,120,130}
\renewcommand{\lstlistingname}{Code}
\makeatletter
\renewcommand{\fnum@lstlisting}{\lstlistingname\nobreakspace\thelstlisting}
\makeatother
\newtcolorbox{codebox}{%
  enhanced,
  colback=codebg,
  colframe=codeborder,
  boxrule=0.5pt,
  arc=3pt,
  boxsep=0pt,
  left=9pt,
  right=9pt,
  top=7pt,
  bottom=7pt,
  before skip=0pt,
  after skip=0pt,
}
\title{Scientific Discovery under Validation Congestion via Multi-Fidelity Pairwise Rankings}

\author{Kevin Tirta Wijaya\textsuperscript{1,2}, Alston Lo\textsuperscript{3}, Michael Sun\textsuperscript{3}, Wojciech Matusik\textsuperscript{3}, Vahid Babaei\textsuperscript{1,2}\\
\textsuperscript{1}University of Bonn, \textsuperscript{2}Fraunhofer SCAI, \textsuperscript{3}Massachusetts Institute of Technology\\
\texttt{kevintirta.w@gmail.com, vbabaei@uni-bonn.de}
}

\usepackage{amsmath,amsfonts,bm}

\newcommand{\methodname}{PRISMS}

\def\eqref#1{equation~\ref{#1}}
\def\1{\bm{1}}

\def\vv{{\bm{v}}}

\DeclareMathAlphabet{\mathsfit}{\encodingdefault}{\sfdefault}{m}{sl}
\SetMathAlphabet{\mathsfit}{bold}{\encodingdefault}{\sfdefault}{bx}{n}

\def\sB{{\mathbb{B}}}

\def\sD{{\mathbb{D}}}
\def\sX{{\mathbb{X}}}

\iclrfinalcopy % Uncomment for camera-ready version, but NOT for submission.
\begin{document}

\maketitle
\begin{abstract}
Modern computational methods can now propose candidate molecules, materials, and other scientific designs at an unprecedented scale, creating a \textit{validation congestion} where candidates are abundant, but experimental capacity to physically evaluate them remains scarce.
Discovering novel scientific designs has therefore become increasingly dependent on \textit{curation}: selecting a small set of promising designs for slow and costly experiments.
Existing curation methods typically rely on data-driven regression models that predict absolute scores, but training these models requires substantial experimental data to begin with.
Yet, useful curation signals do not have to take the form of absolute measurements, as scientific design discovery is often comparative in nature.
Here, we propose that curation can instead be primarily driven by expert \textit{pairwise rankings}, which are substantially easier to gather. 
The expertise can come from computational tools or human input of multiple levels of fidelity, ranging from empirical rules of thumb to agentic workflows and experienced scientists.
We introduce \methodname, a framework that uses pairwise rankings from one or more experts, potentially spanning multiple levels of expertise, to identify the most promising candidates without relying on data-hungry regressors.
When experts differ in fidelity and cost, \methodname\ escalates pairwise queries from lower- to higher-fidelity rankers based on a Fisher-information criterion.
In iterative screening that selects designs from fixed drug discovery libraries, \methodname\ achieves $50\%$ top-10 discovery recall in $\sim\!42\%$ fewer rounds than regression-only active learning, and in $\sim\!15\%$ fewer rounds than the ranking-based method with no selective escalation.
In optimization that generates new designs without restriction to a predefined library, \methodname\ achieves $\sim\!18.8\%$ higher hypervolume than the Bayesian optimization baseline.
These results show that multi-fidelity expert pairwise rankings can provide an efficient alternative to regression-driven curation in scientific design discovery.
\end{abstract}
\section{Introduction}
\begin{figure}[t]
    \centering
    \includegraphics[width=0.95\linewidth]{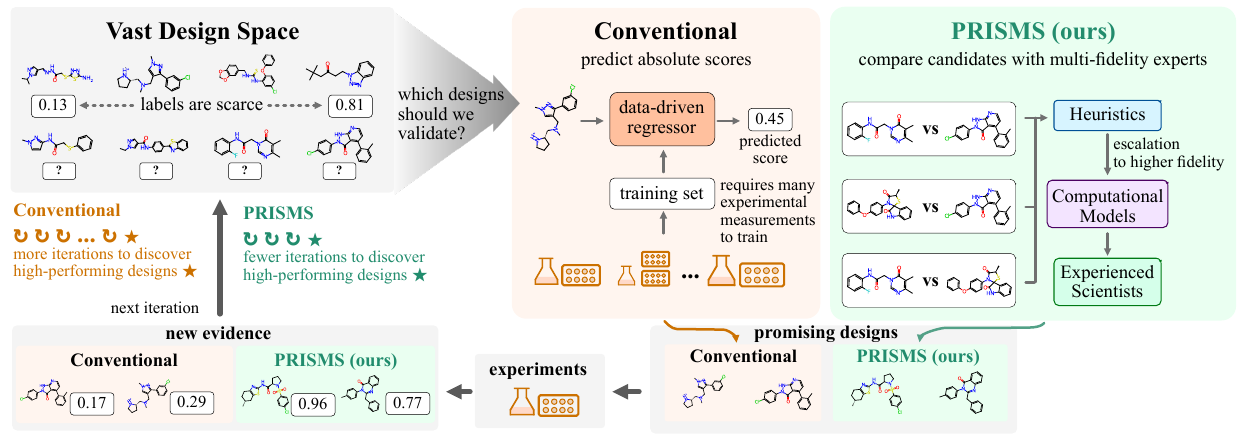}
    \caption{Accelerating scientific discovery through curation. Conventional curation methods predict absolute scores from scarce experimental data, while PRISMS curates designs using selectively queried multi-fidelity pairwise rankers. Better curation sends more promising designs for experiments at each round, reducing the number of design-experiment loops needed for discovery.}
    \label{fig:teaser}
\end{figure}

Modern computational methods can generate or enumerate molecules, materials, and other scientific designs at increasingly unprecedented scales.
Virtual screening libraries, for example, routinely contain $10^5$--$10^9$ compounds \citep{gryniukova2023ai_vsExample,zhou2024artificial_vsExample,luttens2025rapid_vsExample}, while airfoil datasets may contain hundreds of thousands of feasible designs \citep{liu2024afbench_airfoilBenchmark}.
Yet, the physical validation layer, i.e., \textit{making} (synthesis or fabrication) and \textit{testing} (experimental measurement and safety checks) remains slow and costly.
A single experimental round in drug discovery, for example, typically requires one to three months to complete \citep{mas2026back_dmtaCycleTime}.
As a result, scientific design discovery---the iterative process of proposing and validating scientific designs to identify high-performing ones (Figure \ref{fig:teaser})---is increasingly facing \textit{validation congestion}, where many potentially valuable designs compete for limited experimental capacity.
Under validation congestion, the challenge shifts from generating yet another design to allocating scarce experimental resources toward the most promising ones, a process we call \textit{design curation}.

The dominant computational approach for curation is to train a regression model on previously validated designs and select new candidates based on their predicted property values or associated uncertainty estimates \citep{gryniukova2023ai_vsExample, zhou2024artificial_vsExample, luttens2025rapid_vsExample}.
Such approaches can be effective when sufficient training labels (experimental measurements) are available.
Under validation congestion, however, the labels needed to train reliable regression models are themselves expensive to acquire.
This motivates complementary sources of information that can guide curation before sufficient absolute measurements have been collected.

When we reconsider the problem from a broader perspective, scientific design discovery is often comparative in nature. 
The goal is not to predict every design's exact property value, but to identify which designs are better: whether molecule A inhibits a target more strongly than molecule B, whether material A is stronger than material B, or whether airfoil A has a higher lift coefficient than airfoil B.
Useful curation signals, therefore, do not have to take the form of absolute measurements.
They may instead take the form of \textit{pairwise rankings}: comparative judgments of whether one design is more promising than another.
Such rankings can be obtained from heterogeneous sources of scientific knowledge, including empirical rules of thumb \citep{lipinski1997experimental_lipinskiruleof5, mizutani2010hume_alloyRules, george2020limited_crystalRules}, computational models \citep{ORCA_orcadft, trott2010autodock_autodockvina, zeni2025generative_mattergen}, and domain experts, spanning different levels of fidelity and cost.
Inexpensive heuristics may provide abundant but noisy rankings, while experienced scientists may provide more reliable judgments despite being unable to precisely quantify the absolute property values.
Pairwise rankings thus provide a natural interface for integrating heterogeneous scientific knowledge and are often easier to elicit reliably than absolute property estimates \citep{thurstone1927law_rankingIsBetter, stewart2005absolute_rankingIsBetter, routh2023rating_rankingIsBetter, hoeijmakers2024_rankingIsBetter}.

To turn multi-fidelity pairwise rankings into curation decisions, we introduce Pairwise Ranking for Information-guided Scientific curation with Multi-fidelity Sources (\methodname), a graph-based curation framework that aggregates pairwise comparisons across rankers and selectively queries higher-fidelity rankers.
In \methodname, nodes represent candidate designs and store either measured property values or current property estimates, while edges encode pairwise rankings accumulated across rankers and campaign rounds.
The values of unmeasured designs are inferred via a regularized Bradley-Terry model \citep{bradley1952rank}, anchored by available measurements and prior property estimates.
Within each round, \methodname\ queries rankers sequentially in increasing order of fidelity.
Since higher-fidelity rankers often have limited availability, \methodname\ selectively escalates comparisons based on a Fisher information-based criterion \citep{fisher1925theory_finsherinfo}, concentrating higher-fidelity queries on comparisons that can improve the property estimates the most.
The resulting estimates are then used to curate the set of promising designs that merit experimental validation.

We evaluate \methodname\ on 14 drug-discovery and three engineering datasets in two settings: (1) \textit{screening}, which identifies top-performing designs from a fixed library, and (2) \textit{optimization}, which generates new designs and searches for high-performing ones without a predefined library.
In screening, \methodname{} reaches $50\%$ top-10 discovery recall in $\sim\!42\%$ fewer iterations than conventional pool-based active learning.
In optimization, it achieves $\sim\!18.8\%$ higher hypervolume than GP-BO, a Bayesian optimization baseline.
These results demonstrate that multi-fidelity pairwise rankings can effectively guide curation when experimental measurements are scarce both at initialization and throughout the campaign, and accelerate scientific design discovery by reducing the number of required experimental rounds.

\section{Background}
\label{sec:2_related_works}
Scientific discovery may involve either understanding existing systems or identifying new designs with desirable properties \citep{wang2023scientific_ai4science}. 
The former includes discovering scientific relationships, mechanisms, or hypotheses. 
The latter, which we call scientific design discovery and focus on in this paper, concerns identifying new drugs, materials, or engineered objects that satisfy an objective.
Scientific design discovery spans various domains from drug discovery to engineering \citep{reddy2007virtual_VSGeneral, lam2018advances_BOinAeronautics, mathern2021multi_BOinCivilEngineering}. 
As computational candidate design generation becomes increasingly scalable while physical validation remains slow and costly \citep{mas2026back_dmtaCycleTime}, the bottleneck of scientific design discovery shifts to deciding which designs to validate, a process we call curation.

A common curation approach is to use surrogate models to predict candidate designs' property values, reducing reliance on experimental validation. 
When the underlying physics is sufficiently well understood and computationally tractable, physics-based models can provide such surrogates, as in molecular docking \citep{tuccinardi2009docking_dockingVS, kontoyianni2017docking_dockingVS}. 
In many scientific domains, however, such models are unavailable, motivating the widespread use of data-driven surrogates \citep{gryniukova2023ai_vsExample, zhou2024artificial_vsExample, luttens2025rapid_vsExample, ansari2022autoinverse, wijaya2025trustworthy_trustmol}.
For example, in virtual screening, uncertainty-aware surrogates such as Gaussian Processes \citep{williams1995gaussian_gpRegression} can be combined with acquisition functions such as Upper Confidence Bound (UCB) \citep{auer2002finite_ucbAlgo} or Expected Improvement (EI) \citep{jones1998efficient_eiAlgo} to determine which designs to validate next.
These approaches, however, rely primarily on absolute property measurements as supervision, which are themselves scarce under validation congestion.

Pairwise rankings provide an alternative source of supervision.
Rather than requiring the absolute property value of a design, a ranker only determines whether one design is preferable to another.
Pairwise ranking methods can recover latent scores, top-$k$ partitions, or global orderings from such comparisons.
Spectral MLE \citep{chen2015spectral_spectralMLE} estimates global scores under the Bradley-Terry model \citep{bradley1952rank}, while HodgeRank \citep{jiang2011statistical_hodgerank} uses Hodge decomposition to separate globally consistent ranking signals from cyclic inconsistencies.
\citet{mohajer2017active_topKrank} actively select comparisons near the top-$k$ decision boundary to reduce the number of queries needed to identify the top-$k$ partition.
More recently, \citet{qin2024large_prpSlidingK} show that LLMs can serve as pairwise rankers, identifying top-ranked items through sequential pairwise rankings.
For scientific design curation, pairwise rankings can also be used to improve estimates on an absolute property scale, allowing rankings to be combined with experimentally measured values.
Projection \citep{yan2024consolidating_projection} and RankRefine++ \citep{wijayabayesian_rankrefinepp} follow this direction by using pairwise rankings to refine predictions from a learned regressor.
\section{Design Curation with Multi-Fidelity Pairwise Rankings}
%\section{Validation-congested Scientific Discovery}

We consider scientific discovery campaigns under validation congestion, where advances in generative models and growth of large scientific design libraries make candidate designs abundant, while physical validation capacity remains limited.
A campaign covers \textit{multiple} rounds of selecting and experimentally validating a small batch of designs, and the goal of curation is to reduce the number of rounds needed to discover high-performing designs.
Candidate designs come from a fixed library in iterative screening campaigns or are generated throughout an optimization campaign.

\begin{figure}[t]
    \centering
    \includegraphics[width=0.95\linewidth]{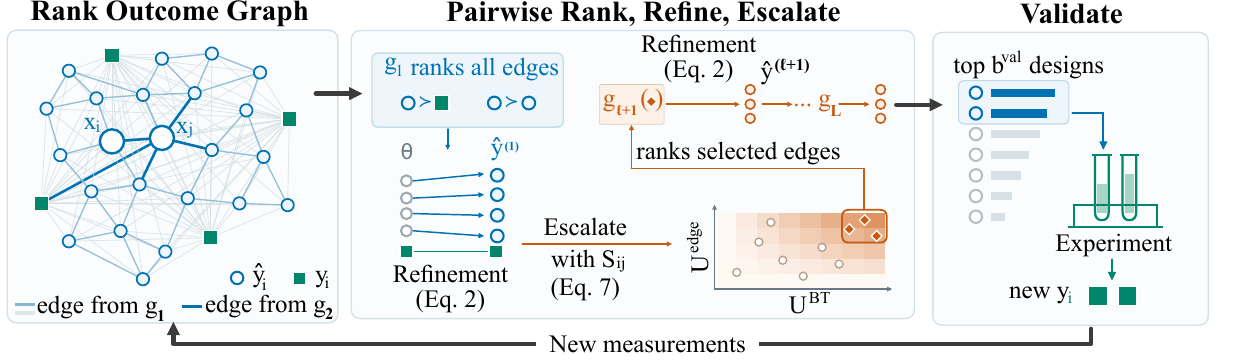}
    \caption{Overview of \methodname. \textbf{(Left)} A rank outcome graph connects unmeasured designs (circles) to all experimentally validated designs (squares) and some unmeasured designs. \textbf{(Middle)} Starting with low-fidelity rankings on all eligible edges, a regularized Bradley-Terry model jointly refines property estimates of all unmeasured designs. Select pairs (edges) are escalated to higher-fidelity rankers based on a Fisher information criterion, and the estimates are further refined after exhausting the budget of a ranker $g_\ell$. \textbf{(Right)} The highest-ranked designs are selected for experimental validation, subject to a budget. New measurements update the rank outcome graph for the next round.}
    \label{fig:prisms_details}
\end{figure}

The overview of \methodname\ is illustrated in Figure~\ref{fig:prisms_details}.
At the start of round $t$, let  $\sX^V_{t-1}$ denote the set of experimentally validated designs from the previous rounds, and $\sX^U_t$ the unmeasured designs available for curation.
Each design $x_i$ has a property value $y_i$ that can be revealed through experimental validation, e.g., a drug candidate's bioactivity measured by an assay.
Given a per-round validation budget $b^\mathrm{val}$, curation selects a small batch $\sB_t~\subseteq~\sX_t^U$ with $\lvert \sB_t \rvert \leq b^\mathrm{val}$.
The experimental validation layer then reveals the property values of the selected designs, and updates the set $\sX^V_t=\sX^V_{t-1}\cup\sB_t$.

In each campaign round, \methodname\ jointly estimates the property values of the unmeasured designs in $\sX^U_t$ based on the measurements of validated designs in $\sX^V_{t-1}$ and the pairwise rankings collected from multi-fidelity pairwise rankers.
The property estimates therefore provide a common interface for combining multi-fidelity pairwise rankings with experimental measurements, which ultimately drives the curation process.

\subsection{Joint Property Estimation from Pairwise Rankings}
We assume access to pairwise rankers $g_1, \dots, g_L$ indexed in increasing level of fidelity.
Each ranker takes two designs as input and predicts which has the higher property value.
Lower-fidelity rankers are assumed to be cheaper to query, whereas higher-fidelity rankers provide more reliable rankings at a limited query budget.
Their judgments may nevertheless be noisy and mutually inconsistent.

Within each campaign round, we query the rankers in $L$ stages of increasing fidelity and store their outcomes in a graph where nodes represent designs and edges represent the pairwise ranking outcomes.
An experimentally validated node carries its ground-truth property value $y_i$, while an unmeasured node carries its estimate $\hat{y}_i$.
At the beginning of each round, we connect each unmeasured design to all experimentally validated designs and the $k^U$ other unmeasured designs closest in prior property estimate $\theta_i$.
The connections remain fixed within a round and are recomputed at the start of the next round.
Only nodes connected by edges are eligible to be ranked by the pairwise rankers.

Omitting the campaign-round index $t$ for clarity, let $\sD^{(\ell)} = \{r^{(\ell)}_{ij}\}$ denote the set of pairwise rankings obtained from $g_\ell$, where $r^{(\ell)}_{ij} \in \{0,1\}$ is the binary outcome and $r^{(\ell)}_{ij} = 1$ indicates that $g_\ell$ judges $x_i$ to have a higher property value than $x_j$.
Let $z_i$ denote the property value assigned to design $x_i$ during the optimization to estimate $\hat{y}_i$. 
Since experimentally validated nodes have their ground-truth values revealed, we keep their values fixed to $z_i=y_i \text{ for all } x_i\in\sX^V$.
Under the Bradley-Terry model \citep{bradley1952rank}, the probability of $x_i \succ x_j$ is
\begin{equation}
p_{ij}(z)=\sigma\left((z_i-z_j) / \tau \right), \qquad \sigma(u)=\frac{1}{1+\exp(-u)},
\end{equation}
where the temperature $\tau$ can be fitted from a small sample of measured designs \citep{wijayabayesian_rankrefinepp}.
At stage $\ell$, we jointly estimate the unmeasured values by minimizing
\begin{equation}
\mathcal L^{(\ell)}(z)
=
\sum_{s=1}^{\ell} \lambda^{(s)}_1
\sum_{r_{ij}^{(s)}\in\sD^{(s)}}
\operatorname{BCE}\left(r_{ij}^{(s)},p_{ij}(z)\right)
+
\lambda_2\sum_{x_i\in\sX^U}(z_i-\theta_i)^2,
\label{eq:joint_estimation}
\end{equation}
where $\operatorname{BCE}(r,p) = -r\log p-(1-r)\log(1-p)$, $\lambda^{(\ell)}_1$ controls the contribution of ranker $g_\ell$, and $\theta_i$ is a prior estimate of the property value with regularization strength $\lambda_2 \geq 0$. 
Minimizing this objective at stage $\ell$ yields an estimate $\hat y_i^{(\ell)}$ for each unmeasured design.
After the final stage $L$, we select $b^\text{val}$ unmeasured designs with the highest estimates $\hat y_i^{(L)}$ for experimental validation.
Note that the prior estimates may be obtained from any estimator that predicts the numerical values of the property, for example, a random forest regressor trained on available measurements at the current round.
Importantly, \methodname\ uses $\theta_i$ only as a weak prior that is subsequently refined using the accumulated pairwise rankings.

\subsection{Information-guided Escalation to Higher-Fidelity Pairwise Ranker}
\label{sec:3_escalation}
We assume that the lowest-fidelity ranker $g_1$ can evaluate all eligible pairs in the comparison graph, producing the rankings $\sD^{(1)}$.
However, higher-fidelity rankers often have limited availability, such that at stage $\ell>1$, ranker $g_\ell$ can evaluate at most $b^{(\ell)}$ pairs.
To allocate this budget, we prioritize comparisons expected to reduce uncertainty in candidate property estimates.
Our escalation criterion combines two measures, \textit{outcome uncertainty} and \textit{edge uncertainty}, defined below.

\paragraph{Outcome Uncertainty.}
Outcome uncertainty is the variance of the binary ranking outcome under the Bradley-Terry model. After stage $\ell$, it is
\begin{equation}
U_{ij}^{\mathrm{BT}}=p_{ij}(\hat y^{(\ell)}) (1-p_{ij}(\hat y^{(\ell)})).
\end{equation}
The outcome uncertainty is maximized when the two designs have equal estimated property values.

\paragraph{Edge Uncertainty.}
Edge uncertainty is the variance of the estimated property difference between two designs, connected by an edge, under a local Gaussian approximation.
We calculate edge uncertainty by using the curvature of the objective in Equation~\ref{eq:joint_estimation} at the current estimates:
\begin{equation}
H^{(\ell)}
=
\left.
\nabla^2\mathcal L^{(\ell)}(z)
\right|_{z=\hat y^{(\ell)}},
\end{equation}
where the Hessian is taken with respect to unmeasured values, holding measured values fixed.
When $\lambda_2>0$, $H^{(\ell)}$ is positive definite, and its inverse gives the covariance of the local Gaussian approximation induced by the objective.

Let $\vv_{ij}$ be an incidence vector restricted to the unmeasured variables, containing $+1$ for $x_i$ if $x_i~\in~\sX^U$, $-1$ for $x_j$ if $x_j~\in~\sX^U$, and zeros elsewhere.
We define edge uncertainty as the variance
of the property difference under the approximation,
\begin{equation}
U_{ij}^{\mathrm{edge}}
=
\vv_{ij}^{\mathsf T}(H^{(\ell)})^{-1}\vv_{ij}.
\end{equation}
High edge uncertainty indicates that the property difference of a pair is poorly constrained under the local approximation, making the pair a potential target for more accurate ranking information.

\paragraph{Information-guided Escalation.}
For $\ell < L$, we evaluate the contribution of a new ranking from $g_{\ell+1}$ for pair $(x_i, x_j)$ at the current estimates $\hat{y}^{(\ell)}$. 
The additional weighted Bradley-Terry term yields the following rank-one update to the local precision:
\begin{equation}
H'
=
H^{(\ell)}
+
\frac{\lambda_1^{(\ell+1)}}{\tau^2}
U_{ij}^{\mathrm{BT}}\vv_{ij}\vv_{ij}^{\mathsf T}.    
\end{equation}
By the matrix determinant lemma, the resulting increase
in log-determinant is
\begin{equation}
\begin{aligned}
S_{ij}
&=\log\det H'-\log\det H^{(\ell)}
\\
&=\log\!\left(
1+ \frac{\lambda_1^{(\ell+1)}}{\tau^2}
U_{ij}^{\mathrm{BT}}U_{ij}^{\mathrm{edge}}
\right).
\end{aligned}
\label{eq:escalation_score}
\end{equation}
This local information-gain score, derived from the Bradley-Terry Fisher information, favors pairs with both high outcome uncertainty and high edge uncertainty.
We select up to $b^{(\ell+1)}$ eligible pairs with the highest individual information-gain scores, all evaluated using the current precision matrix, and query ranker $g_{\ell+1}$ on these pairs.
The returned rankings form $\sD^{(\ell+1)}$, and the property estimates are updated by minimizing Equation~\ref{eq:joint_estimation}.
This process continues up to the last ranker at stage $L$, after which the final estimates are used to determine which designs are sent to the experimental validation layer.

\subsection{RankRefine++ as a Special Case}
\label{sec:3_rankrefine++}
RankRefine++ \citep{wijayabayesian_rankrefinepp}, the closest work to \methodname, refines a pretrained regressor's predictions using pairwise comparisons against experimentally validated designs.
Its per-design estimation objective is recovered as a
special case of Equation \ref{eq:joint_estimation} by considering only a single unmeasured design $x_0$, connected only to measured designs, and using only a single ranker $g_1$.

Let $\theta_0$ be the regressor's prediction and
$\sigma_{\mathrm{re}}^2$ its associated uncertainty.
For $\lambda_1^{(1)}>0$, setting
$
\frac{\lambda_2}{\lambda_1^{(1)}}
=
\frac{1}{2\sigma_{\mathrm{re}}^2}
$
makes Equation~\ref{eq:joint_estimation} proportional to the per-design MAP objective of RankRefine++, yielding the same estimate.
\section{Experimental Results}

% Scientific rediscovery campaign.
\subsection{Drug Discovery Campaign}
\label{sec:main_drug_campaign}
\paragraph{Setup.}
We evaluate \methodname\ and competing curation methods in simulated iterative screening campaigns that aim to identify the top-$10$ molecules in a fixed library using as few experimental rounds as possible.
We use $10$ ADMET tasks \citep{ainsley_2026_21504733_openadmetExpansionrx, macdermott_opeskin_2026_21987121_openadmetCYP} and $4$ binding-affinity tasks \citep{gorantla2024benchmarking_dockingDataset}, detailed in Appendix Table~\ref{tab:dataset}, and each experiment is repeated over $30$ random seeds.
For each task-seed pair $u$, where $n_\text{mol}^{(u)}$ is the size of the molecular dataset used in $u$, we construct a library of $\min(n_\text{mol}^{(u)},1010)$ randomly sampled molecules, reveal the labels of $10$ molecules to initialize $\sX^V_0$, and treat the remainder as unlabeled.
At each round, the curation selects $10$ unlabeled molecules for experimental validation, simulated by revealing their labels.

%\paragraph{Pairwise Rankers.}
Unless otherwise stated, we use two pairwise rankers of different fidelities.
The lower-fidelity ranker $g_1$ is an ensemble of a rule-based classifier written by Claude Sonnet 4.6, a random forest \citep{breiman2001random_randomforest}, and a support vector machine \citep{cortes1995support_svm}; at each round $t$, the latter two are refitted on pairwise rankings derived from $\sX^V_{t-1}$.
On a separate 24-round drug-discovery campaign evaluation, $g_1$'s pairwise accuracy increases from approximately $60\%$ in round 1 to $68\%$ by round 9 and remains stable afterwards.
The higher-fidelity ranker $g_2$ is a simulated $75\%$-accurate oracle, implemented by flipping ground-truth rankings with probability $25\%$, with a budget of $b^{(2)} = 100$ queries per round.
We initialize $\theta_i$ using a random forest regressor to build the initial unlabeled-unlabeled node connections, and update $\theta_i$ with a Bradley–Terry estimate obtained from $g_1$ rankings between the query and labeled nodes.

%\paragraph{Baselines and Metrics.}
We compare \methodname{} with seven baselines described in Section~\ref{sec:2_related_works}: random sampling, a regression-based Gaussian Process with Expected Improvement algorithm (GP-EI), a pairwise-ranking based GP-EI, HodgeRank, Active Top-k, Spectral MLE, and RankRefine++.
All baselines except random sampling and GP-EI have access to the same multi-fidelity rankers and per-round query budgets as \methodname. 
However, they allocate higher-fidelity queries to randomly selected pairs, without information-based escalation.
We evaluate discovery efficiency and ranking quality using four metrics:
(1) \textit{discovery recall}: the fraction of true top-10 molecules cumulatively discovered at round $t$;
(2) \textit{NDCG}: the normalized discounted cumulative gain over the current candidate pool; 
(3) \textit{top-$k\%$ F1}, the F1 score for identifying molecules belonging to the true top-$k\%$ of the library; and
(4) \textit{restricted mean time to 50\% discovery recall}: the area under the Kaplan-Meier survival curve \citep{kaplan1958nonparametric_kmrmst} for the time to discover five of the true top-10 molecules.
We report discovery and ranking performance over the first 12 rounds, corresponding to approximately 1--3 years of experiments under the assumed duration of 1--3 months per round \citep{mas2026back_dmtaCycleTime}. 
The full 24-round campaigns are used only to estimate the restricted mean time to 50\% discovery recall.

\begin{figure}[ht]
    \centering
    \setlength{\abovecaptionskip}{4pt}
    \includegraphics[width=0.95\linewidth]{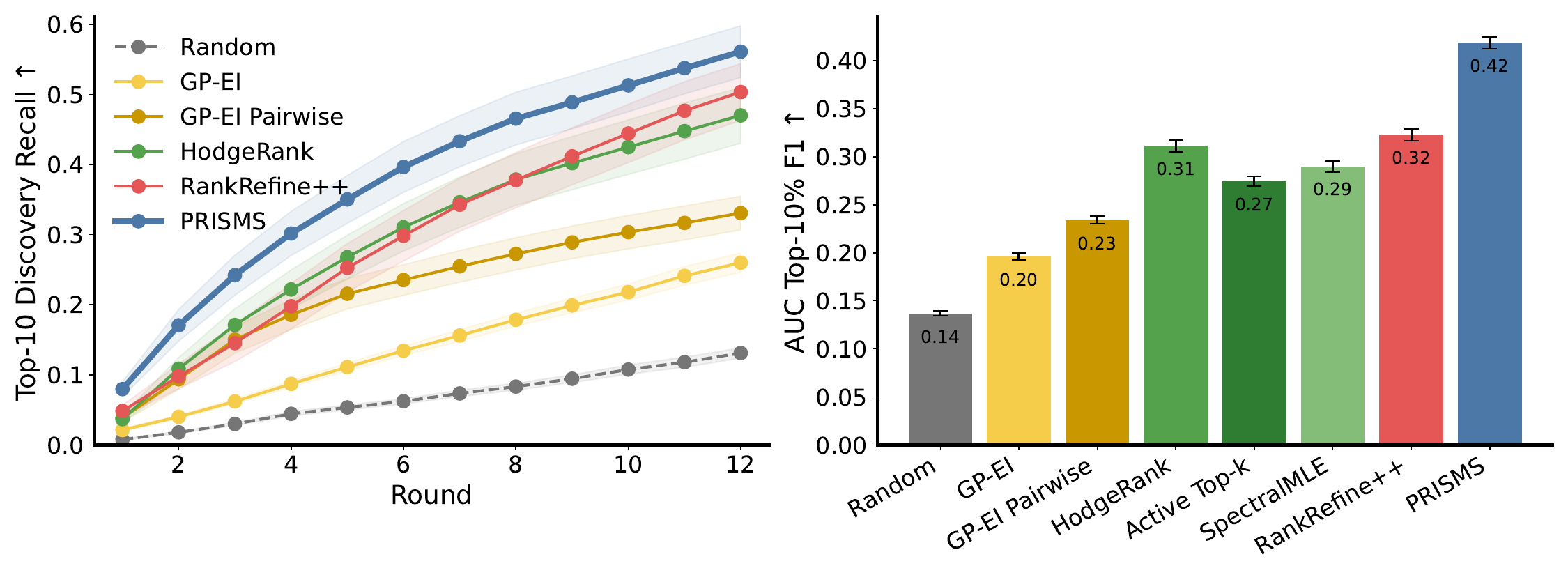}
    \caption{Aggregated results of 12-round scientific design discovery campaigns on 14 drug discovery datasets (ADMET and binding affinity). We report the mean and standard error over 30 random seeds. \textbf{Left.} Top-10 discovery recall over 12 rounds. \textbf{Right.} AUC of Top-10\% F1 score.}
    \label{fig:main_screening}
\end{figure}

\paragraph{Results.}
Figure \ref{fig:main_screening} summarizes the 12-round iterative screening campaigns across datasets and runs.
\methodname\ achieves the highest top-10 discovery recall curve throughout all rounds, improving area under the top-10 discovery recall curve (AUDC@10) by $25.1\%$ over HodgeRank and $24.3\%$ over RankRefine++.
\methodname\ also improves the top-$10\%$ F1 score, where \methodname\ outperforms RankRefine++ by $31.2\%$. 
Because RankRefine++ is a special case of \methodname\ (Section~\ref{sec:3_rankrefine++}), the gains indicate the benefits of connecting unlabeled nodes, joint optimization, and information-based escalation under equal access to multi-fidelity pairwise rankers.
More importantly, \methodname\ achieves a 24-round restricted mean time to 50\% discovery recall of $11.1$ rounds, compared with conventional Bayesian approaches such as GP-EI ($19.3$ rounds) that do not use pairwise rankers.
Under validation congestion, assuming 1--3 months per experimental round, this translates to an estimated 8--24 months of experimental cycle time saved.
Finally, GP-EI Pairwise achieves $\sim\!61\%$ higher AUDC@10 than GP-EI, highlighting the value of pairwise ranking supervisions when numerical labels are scarce.
Detailed results are shown in Table \ref{tab:full_main_auc_table}, along with a scaling analysis in Section~\ref{sec:scaling} and a proof of concept for real-world applications with non-simulated rankers is shown in Appendix Section~\ref{sec:agentic_prisms}.

\subsection{Engineering Discovery Campaign}
\paragraph{Setup.}
We evaluate \methodname\ for iterative screening beyond drug discovery on three engineering datasets: 3D-printed structures (mechanical toughness) \citep{gongora2020bayesian_crossBarrelDataset}, dye laser configurations (peak score), and battery redoxmers (absorption wavelength deviation) \citep{hase_olympus_2021_olympusBench}.
We use $10$ random seeds per dataset and otherwise follow the setup in Section~\ref{sec:main_drug_campaign}.

\paragraph{Results.}
Figure \ref{fig:engineering_screening} in the Appendix shows the aggregated results of the 12-round campaigns for the engineering datasets. 
Consistent with the findings in the drug discovery domain, \methodname\ outperforms all evaluated baselines across all performance metrics.

\subsection{Multi-Objective Drug Optimization Campaign}
\begin{wrapfigure}{r}{0.45\linewidth}
    \centering
    \setlength{\abovecaptionskip}{4pt}
    \includegraphics[width=0.99\linewidth]{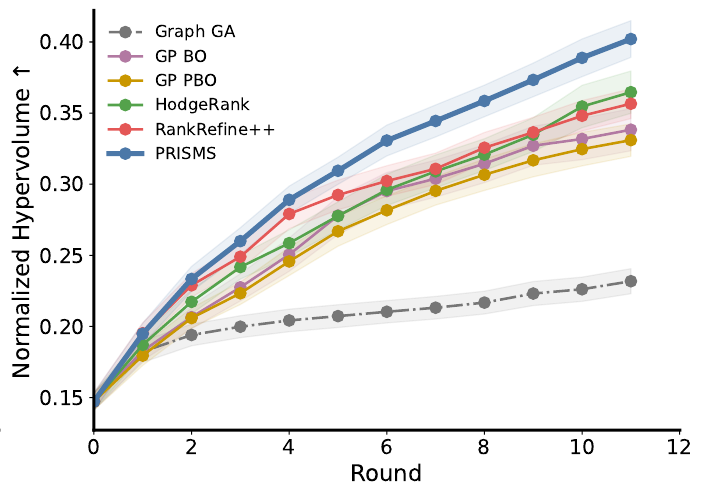}
    \caption{Normalized hypervolume over 12 campaign rounds and 30 random seeds for multi-objective optimization targeting JNK3 activity and solubility}
    \label{fig:moo}
\end{wrapfigure}
\paragraph{Setup.}
Beyond screening, we evaluate \methodname\ in multi-objective molecular optimization targeting oracle-evaluated JNK3 activity \citep{li2018multi} and solubility \citep{swanson2024admet_admetai}.
Unlike fixed-library screening, candidates are generated throughout the campaign using molecular graph-based crossover and mutation operators \citep{jensen2019graph_graphGA} in a genetic algorithm-style loop.
Each curation method independently predicts both objectives for newly generated molecules to select the promising candidates.

We initialize each campaign with 10 randomly sampled molecules from ZINC 250K \citep{sterling2015zinc, gomez2018automatic_zinc} and allow 10 designs per round for oracle evaluation.
The baselines are RankRefine++, HodgeRank, Graph GA \citep{jensen2019graph_graphGA}, GP-BO \citep{tripp2024diagnosing_gpbo} and its preferential Bayesian optimization \citep{gonzalez2017preferential_pbo} extension, GP-PBO.
Performance is measured using the hypervolume of the oracle-validated designs, normalized to the maximum in ZINC250K.

\paragraph{Results.}
Figure \ref{fig:moo} shows the normalized hypervolume across the campaign rounds.
\methodname\ consistently outperforms all other baselines, especially in later rounds when pairwise evidences have accumulated.
At the end of a 12-round campaign, \methodname\ achieves about $14.3\%$ higher hypervolume than HodgeRank and RankRefine++, suggesting that pairwise-ranking-based curation remains effective even when the candidate design set grows over time.

\subsection{Ablation Study}
\paragraph{Setup.}
We ablate the main components of \methodname\ using a similar setup as described in Section~\ref{sec:main_drug_campaign}, run for 24 rounds. 
Starting from the default two-fidelity configuration, we separately remove graph edges between two unlabeled nodes, edge uncertainty, and outcome uncertainty; replace information-guided escalation with random escalation; and randomize the Stage-1 initialization. 
We additionally vary the number of fidelity levels, comparing one-, two-, and three-fidelity configurations. 
For the three-fidelity setup, we set $g_3$ to $90\%$ accuracy with a budget $b^{(3)} = 50$.

\paragraph{Results.}
Table~\ref{tab:full_ablation} in the Appendix shows that \methodname' discovery gains come from both access to additional pairwise rankings and how those rankings are propagated and acquired.
Removing graph edges between unlabeled nodes, which prevents ranking information from propagating directly between candidate designs, reduces AUDC@10 by $7.6\%$.
Within the escalation criterion (Section~\ref{sec:3_escalation}), removing edge or outcome uncertainty decreases AUDC@10 by $4.6\%$ and $2.3\%$, respectively, while replacing both with random escalation yields a $7.7\%$ reduction, showing that the two uncertainty terms provide complementary signals for allocating higher-fidelity queries.
The number of fidelity stages has the largest effect: restricting \methodname\ to a single stage lowers AUDC@10 by $16.5\%$, whereas introducing a third fidelity stage improves AUDC@10 by $11.4\%$ over the default two-stage configuration.
These results suggest that, under validation congestion, effectively exploiting external sources of knowledge can substantially improve discovery performance.

\subsection{Sensitivity Analysis}
\paragraph{Setup.}
We study the sensitivity of \methodname\ to the accuracy, availability, and error structure of the higher-fidelity ranker, in a similar setup as described in Section~\ref{sec:main_drug_campaign}.
The pairwise accuracy and per-round query budget $b^{(2)}$ of $g_2$ are varied from $60\%$ to $90\%$ and from $10$ to $200$, respectively.
We consider three noise structures: \textit{uniform flip noise}, which flips pairwise outcomes with uniform probability; \textit{difficulty-dependent noise}, where error probability follows the sigmoid of the difference in property values between the compared designs (designs with similar values are harder to rank accurately); and \textit{structured noise}, which introduces scaffold-dependent biases.
We report \methodname' AUDC@10 and its difference to RankRefine++'s under matched ranker settings and budgets.

\begin{figure}[ht]
    \centering
    \includegraphics[width=0.95\linewidth]{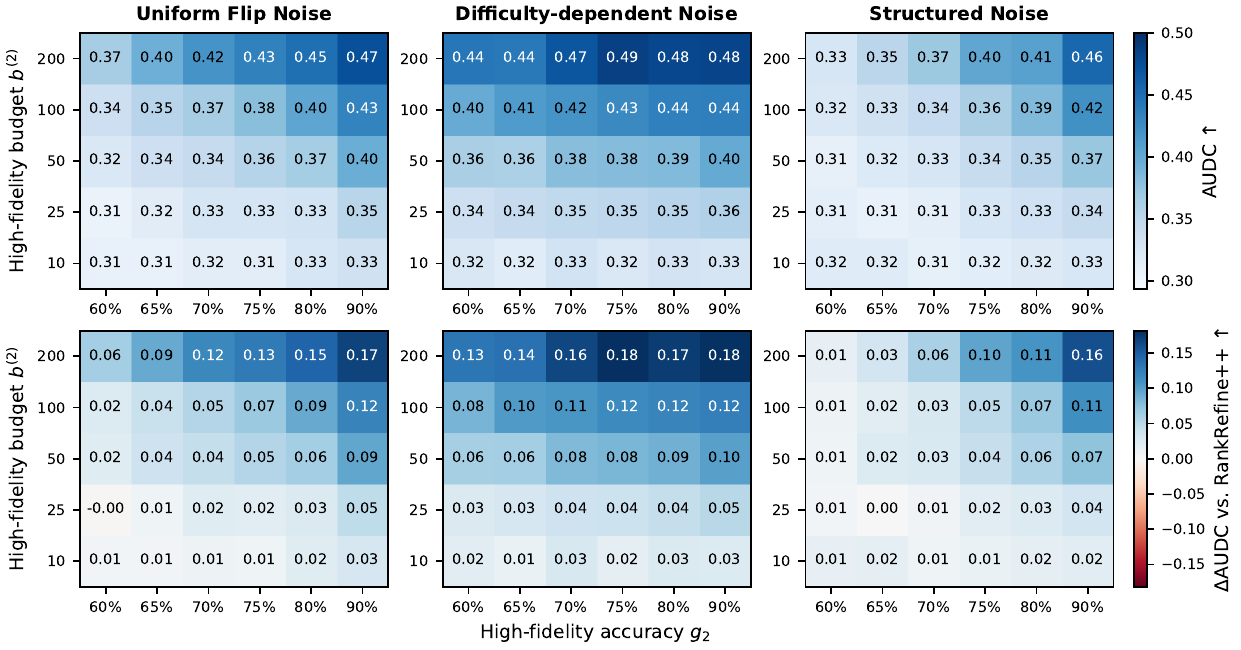}
    \caption{Sensitivity to higher-fidelity ranker accuracy and query budget under three ranking-noise structures. \textbf{(Top)}: \methodname' AUDC@10. \textbf{(Bottom)}: AUDC@10 differences between \methodname\ and RankRefine++ under matched settings.}
    \label{fig:sensitivity}
\end{figure}

\paragraph{Results.}
Figure~\ref{fig:sensitivity} shows that \methodname\ increasing the higher-fidelity query budget consistently improves AUDC@10 across all three noise models.
Increasing ranker accuracy has a modest effect when the query budget is small, but yields increasingly larger gains as more higher-fidelity rankings become available.
\methodname\ also achieves higher AUDC@10 than RankRefine++ in nearly all evaluated settings, with differences generally widening as the budget increases.
This indicates that PRISMS makes more effective use of additional ranking information than RankRefine++
Difficulty-dependent noise yields the strongest overall performance and largest gains over RankRefine++, supporting the relevance of Agentic \methodname\ with an LLM as a ranker, whose errors exhibit a similar structure as discussed in Appendix Section~\ref{sec:agentic_prisms}.
More results for sensitivity analysis are available in Appendix~\ref{sec:app_sensitivity}.

\subsection{Recovery of Missed Top-10 Designs}

\paragraph{Setup.}
We further examine the capability to recover high-performing designs that are not selected for experimental validation in earlier iterative screening rounds. 
Such early-round failures may arise especially when the initial labels or pairwise rankers are biased.
We quantify the recovery using the Earth Mover's Distance (EMD) \citep{rubner1998metric_emd}, which measures how far the true top-$k$ designs are from occupying the top-$k$ positions at round $t$.
For a fixed library of size $N$, let $r_1 < \cdots < r_k$ denote the positions occupied by the true top-$k$ designs in the predicted ranking.
The target is for these designs to occupy positions $1,\ldots,k$ in any order.
We define the normalized EMD as
\begin{equation}
\mathrm{EMD}_{\text{norm}}@k
=
\frac{1}{k(N-k)}
\sum_{i=1}^{k}(r_i-i).
\end{equation}

\begin{figure}
    \centering
    \includegraphics[width=0.95\linewidth]{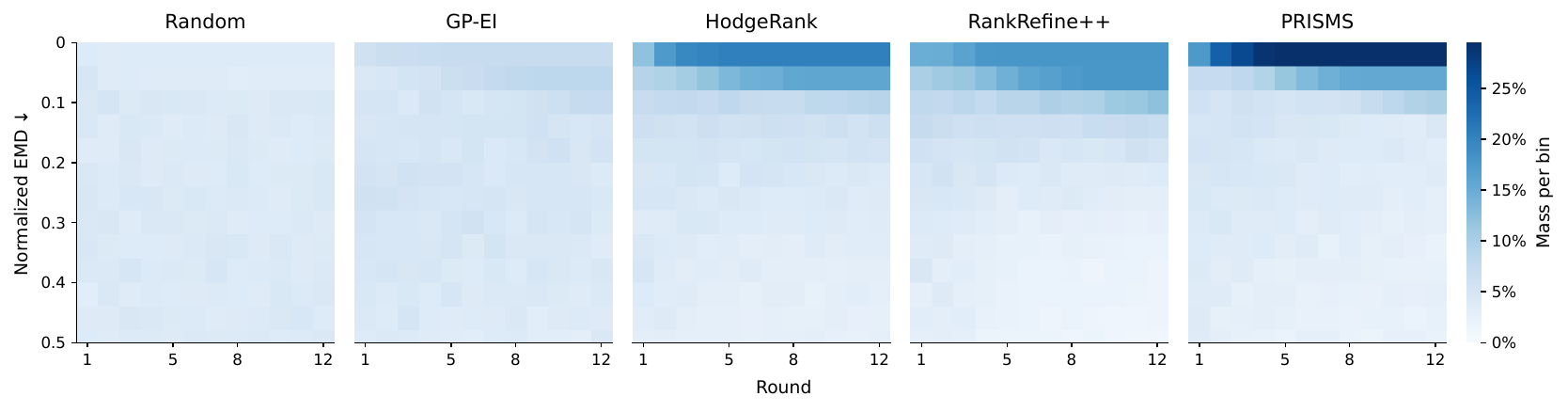}
    \caption{Recovery of true top-10 designs across campaign rounds. Heatmaps show the distribution of normalized Earth Mover's Distance \citep{rubner1998metric_emd} for top-10 designs ($\mathrm{EMD}_\text{norm}@10$, lower is better) over 12 rounds across 14 drug-discovery datasets and 30 random seeds. An $\mathrm{EMD}_\text{norm}@10 = 0.0$ places all true top-10 designs in the top-10 positions, whereas $\mathrm{EMD}_\text{norm}@10 = 1.0$ places them in the bottom-10 positions. Across rounds, \methodname\ shifts true top-10 designs toward the highest ranks faster than competing methods.}
    \label{fig:emd_distribution}
\end{figure}

\paragraph{Results.}
Figure \ref{fig:emd_distribution} shows that \methodname\ progressively shifts true top-10 designs toward top-$10$ positions more quickly than competing methods, as suggested by the growing concentration of samples near $\mathrm{EMD}_\text{norm}@10 = 0$. 
GP-EI remains close to random sampling, while HodgeRank and RankRefine++ reduce $\mathrm{EMD}_\text{norm}@10$ but retain more spread distributions. 
Overall, \methodname\ outperforms other methods at keeping missed high-performing designs near the top of the ranking, increasing their likelihood of selection in subsequent rounds.

\newpage
\section{Conclusion}
Scientific design discovery increasingly faces \textit{validation congestion}, where candidate designs are abundant but physical validation capacity remains scarce.
We have introduced \methodname\ to address this bottleneck by shifting \textit{design curation} away from data-hungry regression toward multi-fidelity pairwise rankings, which provide useful supervision when experimental measurements are scarce and are often easier to elicit reliably.
\methodname\ leverages heterogeneous sources of scientific knowledge by integrating rankings from diverse experts and selectively escalating to higher-fidelity rankers.
Across fixed-library screening and optimization, this approach improves both the rate and extent of discovery of high-performing designs.
More broadly, our results suggest that accelerating scientific design discovery under validation congestion does not have to rely solely on acquiring more measurements, but can also come from making better use of external scientific knowledge that is already available.
\subsection*{AI use statement}
In this work, we used generative AI tools to assist us in implementing functions, identifying relevant literature, and improving readability of sentences.
We have not used generative AI tools for generating synthetic data sets, helping develop theoretical models or conceptual frameworks, formulating mathematical claims, providing critical ingredients for proving mathematical claims, proposing or refining hypotheses, designing or providing feedback on research  methodology or experiments, assisting translation, cleaning or reformatting datasets, supporting qualitative and thematic data analysis, or interpreting results.
We have reviewed and verified all AI-assisted work including LLM-generated codes and text edits.

\bibliography{iclr2027_conference}
\bibliographystyle{iclr2027_conference}

\newpage
\appendix
\section{Limitations}
Our study has several limitations that motivate future works:
\begin{itemize}
    \item Most experiments in the paper are retrospective, using past experimental data, while the higher-fidelity rankers are primarily simulated using noisy oracles.
    This controlled setting allows us to compare curation methods under matched access to ranking information and systematically examine their sensitivity to ranker accuracy and error structure.
    This setting is therefore appropriate for comparing the performance of different curation methods.
    Real-world campaigns may nevertheless introduce challenges beyond those captured by our evaluation, including distribution shifts and domain-specific biases not represented by the tested noise models.

    \item Our agentic-ranker serves as a proof of concept that considers a single LLM-based ranker across three drug-discovery tasks and three random seeds.
    We designed this experiment as an exploratory study, motivated by the accessibility of LLMs and their potential to provide pairwise rankings from their broad knowledge.
    A systematic benchmark study across multiple LLMs would provide complementary insights about their reliability and suitability as ranking sources.
    Such evaluations must also account for task-specific restrictions: in our setup, Claude Fable declined biochemical ranking queries under its safety mechanisms.
    
    \item \methodname\ assumes that rankers have a fixed ordering of fidelity, with higher-fidelity sources expected to provide more reliable rankings.
    This assumption allows us to study selective escalation in isolation from ranker reliability.
    In practice, ranker quality may depend on the selected pairs, scientific domain, and experimental measurements accumulated during the campaign.
    For example, an expert chemist may rank molecules more accurately within familiar scaffolds than unfamiliar ones.
    As experimental evidence accumulates, their accuracy on initially unfamiliar scaffolds may also improve.
    Extending \methodname\ to accommodate such variations during a discovery campaign is an important direction for future work.
\end{itemize}

\section{More Experimental Results}
\subsection{Computational Scaling Analysis}
\label{sec:scaling}
\paragraph{Setup.}
We evaluate the computational scalability of \methodname\ on synthetic design libraries with sizes ranging from $10^3$ to $10^5$ candidates.
For each library size, we measure the 1-round wall-clock runtime and peak resident set size (RSS) during execution of \methodname\ with two levels of ranker fidelity.
Wall-clock runtime is measured on Intel Xeon Sapphire Rapids 48-core/96-thread 2.10GHz.

\paragraph{Results.}
\methodname\ scales to libraries of up to $N=100{,}000$ with moderate memory growth.
Wall-clock runtime and peak RSS increase from $1.45\,\text{s}$ and $282\,\text{MB}$ at $N=1{,}000$ to $32.47\,\text{s}$ and $356\,\text{MB}$ at $N=10{,}000$, and to $3{,}233\,\text{s}$ and $1.03\,\text{GB}$ at $N=100{,}000$.
Even at the largest scale, the computational overhead is small relative to the physical validation layer, which requires days or months to complete, and is therefore unlikely to be the dominant bottleneck.

\subsection{Sensitivity Analysis}
\label{sec:app_sensitivity}
\begin{figure}[ht]
    \centering
    \includegraphics[width=0.85\linewidth]{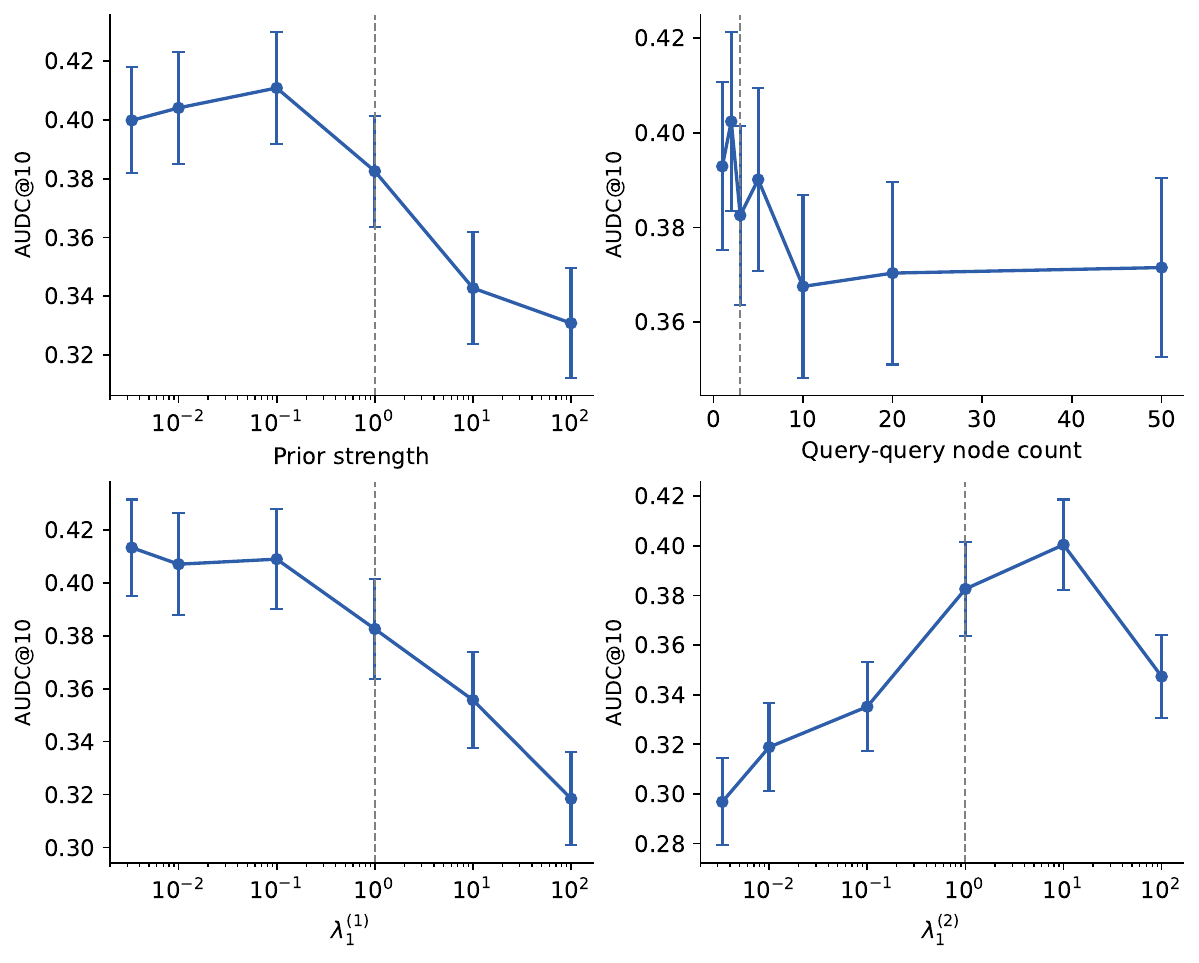}
    \caption{Sensitivity of \methodname{} to prior strength $\lambda_2$, query-query neighborhood size $k^U$, and low- and high-fidelity ranker weights $\lambda_1^{(1)}$ and $\lambda_1^{(2)}$. Each parameter is varied while the others remain at their default values, indicated by vertical dashed lines. Points show mean AUDC@10.}
    \label{fig:parameter_sensitivity}
\end{figure}
Figure~\ref{fig:parameter_sensitivity} shows the effects of prior strength $\lambda_2$, query-query neighborhood size $k^U$, and
ranker weights $\lambda_1^{(1)}$ and $\lambda_1^{(2)}$ to \methodname' performance.
The AUDC@10 is relatively stable when the prior regularization is weak ($\lambda_2 \leq 10^{-1}$), but declines as regularization becomes stronger, reaching approximately $0.33$ AUDC@10 at $\lambda_2=100$.
This pattern supports using regressor predictions as weak priors that pairwise rankings can refine in the label-scarce regimes.

Increasing the low-fidelity weight $\lambda_1^{(1)}$ beyond $10^{-1}$ reduces AUDC@10 from approximately $0.41$ to $0.32$. 
In contrast, increasing the high-fidelity weight
$\lambda_1^{(2)}$ initially improves performance, with the highest observed mean near $\lambda_1^{(2)}=10$, before performance declines at $100$. 
These results support assigning greater weights, but not full confidence, in higher-fidelity rankers with higher accuracy.

Smaller query-query neighborhoods $k^U$ generally yield higher AUDC@10 than larger neighborhoods, with performance approximately plateauing for $k^U \geq 10$. 
Thus, denser connectivity does not necessarily
improve discovery. 

\subsection{Agentic \methodname\ with LLM as Higher-Fidelity Ranker}
\label{sec:agentic_prisms}
\paragraph{Setup.}
We investigate whether \methodname\ can operate with an LLM-based agent as the higher-fidelity ranker, replacing the simulated oracle used in the previous experiments. 
We evaluate three drug-discovery tasks \citep{ainsley_2026_21504733_openadmetExpansionrx, macdermott_opeskin_2026_21987121_openadmetCYP, gorantla2024benchmarking_dockingDataset} targeting kinetic solubility, CYP inhibition, and dopamine D2 receptor activity. 
Each campaign runs for 12 rounds and is repeated over three random seeds. 
For the agentic setting, we use Claude Opus 5.5 as the higher-fidelity pairwise ranker, keeping the ensemble classifiers as the lower-fidelity ranker. 

\paragraph{Results.}
Agentic \methodname\ achieves an area under the discovery recall curve (AUC) of $0.105$, compared with $0.162$ for \methodname\ using the $75\%$-accurate simulated oracle. 
Despite the gap to the oracle setting, Agentic \methodname\ outperforms the oracle-based RankRefine++ and HodgeRank baselines, which achieve AUCs of $0.093$ and $0.092$, respectively. 
These results suggest the feasibility of Agentic \methodname\ for real-world scientific design curation.

\subsection{Characterization of LLM as a Pairwise Ranker}
\paragraph{Setup.}
We characterize the error structure of the LLM-based higher-fidelity ranker (Claude Opus 5.5 High effort) by examining how its pairwise accuracy varies with the difficulty of a comparison.
For each ranked pair $(x_i,x_j)$, we define the standardized property gap as
$d_{ij}=|y_i-y_j|/s_y$, where $s_y$ is the standard deviation of the property values in the corresponding dataset.
We bin pairwise comparisons by $d_{ij}$ and compute the empirical pairwise ranking accuracy within each bin.
We report results separately for the D2R, kinetic-solubility, and CYP-inhibition datasets, along with the pooled accuracy across all three datasets and three random seeds.

\begin{figure}[ht]
    \centering
    \includegraphics[width=0.6\linewidth]{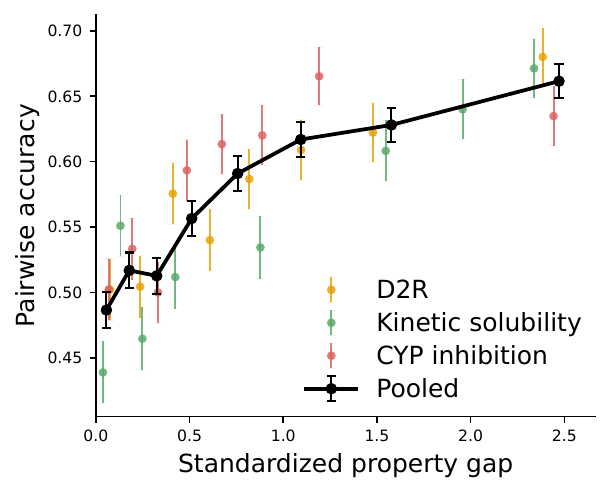}
    \caption{
    Pairwise ranking accuracy of the LLM-based ranker as a function of the standardized true property gap $d_{ij}=|y_i-y_j|/s_y$, where $s_y$ is the dataset-level standard deviation of the property values. 
    Accuracy generally increases with property gap, indicating a difficulty-dependent error structure in which pairs with similar property values are more difficult for the LLM to rank correctly.
    }
    \label{fig:llm_characterization}
\end{figure}
\paragraph{Results.}
Figure~\ref{fig:llm_characterization} shows a dependence of LLM ranking accuracy on the standardized property gap.
For pairs with similar property values, the LLM operates near chance level, whereas its accuracy generally increases as the property difference grows, reaching approximately 0.65--0.66 for the largest gaps in the pooled results.
The same trend is visible across the individual datasets.
These results indicate that LLM ranking errors are \textit{difficulty-dependent}, where comparisons between similar candidates are substantially harder than comparisons between candidates with clearly separated property values.

\subsection{Drug Discovery Campaign}

\begin{table}[ht]
    \centering
    \caption{Experimental results for iterative screening in drug discovery datasets, repeated over 30 random seeds.}
        \begin{tabular}{lccccc}
            \toprule
            Method & \multicolumn{4}{c}{12-round AUC} & \multicolumn{1}{c}{24-round RMST} \\
            \cmidrule(lr){2-5}
             & Recall & NDCG & Top-1\% F1 & Top-5\% F1 & to 50\% Recall \\
            \cmidrule(lr){1-6}
            Random Sampling & 0.077 {\scriptsize(0.005)} & 0.334 {\scriptsize(0.057)} & 0.120 {\scriptsize(0.015)} & 0.134 {\scriptsize(0.005)} & 23.3 \\
            \cmidrule(lr){1-6}
            GP-EI & 0.141 {\scriptsize(0.015)} & 0.398 {\scriptsize(0.057)} & 0.192 {\scriptsize(0.027)} & 0.206 {\scriptsize(0.010)} & 19.3 \\
            GP-EI Pairwise & 0.227 {\scriptsize(0.026)} & 0.416 {\scriptsize(0.067)} & 0.313 {\scriptsize(0.043)} & 0.260 {\scriptsize(0.013)} & 17.9\\
            \cmidrule(lr){1-6}
            HodgeRank  & 0.306 {\scriptsize(0.042)} & 0.490 {\scriptsize(0.055)} & 0.372 {\scriptsize(0.055)} & 0.326 {\scriptsize(0.024)} & 13.5\\
            Active top-k & 0.306 {\scriptsize(0.043)} & 0.487 {\scriptsize(0.055)} & 0.248 {\scriptsize(0.022)} & 0.307 {\scriptsize(0.017)} & 13.6 \\
            Spectral MLE & 0.292 {\scriptsize(0.046)} & 0.494 {\scriptsize(0.056)} & 0.354 {\scriptsize(0.059)} & 0.312 {\scriptsize(0.022)} & 13.7\\
            \cmidrule(lr){1-6}
            RankRefine++ & 0.308 {\scriptsize(0.053)} & 0.489 {\scriptsize(0.052)} & 0.353 {\scriptsize(0.061)} & 0.326 {\scriptsize(0.029)} & 13.1\\
            PRISMS & 0.383 {\scriptsize(0.050)} & 0.541 {\scriptsize(0.054)} & 0.421 {\scriptsize(0.051)} & 0.421 {\scriptsize(0.024)} & 11.1\\
            \bottomrule
        \end{tabular}
    \label{tab:full_main_auc_table}
\end{table}
Table~\ref{tab:full_main_auc_table} shows the detailed results for the iterative screening campaigns in drug discovery described in Section~\ref{sec:main_drug_campaign}. 
The AUC metrics summarize the first 12 rounds. 
To estimate the restricted mean time to 50\% discovery recall (RMST), all methods were evaluated for 24 rounds. 
We calculate RMST as the area under the Kaplan–Meier survival curve from round 0 to round 24, with campaigns that have not reached 50\% recall by round 24 censored at that horizon.

\begin{figure}[ht]
    \centering
    \includegraphics[width=0.95\linewidth]{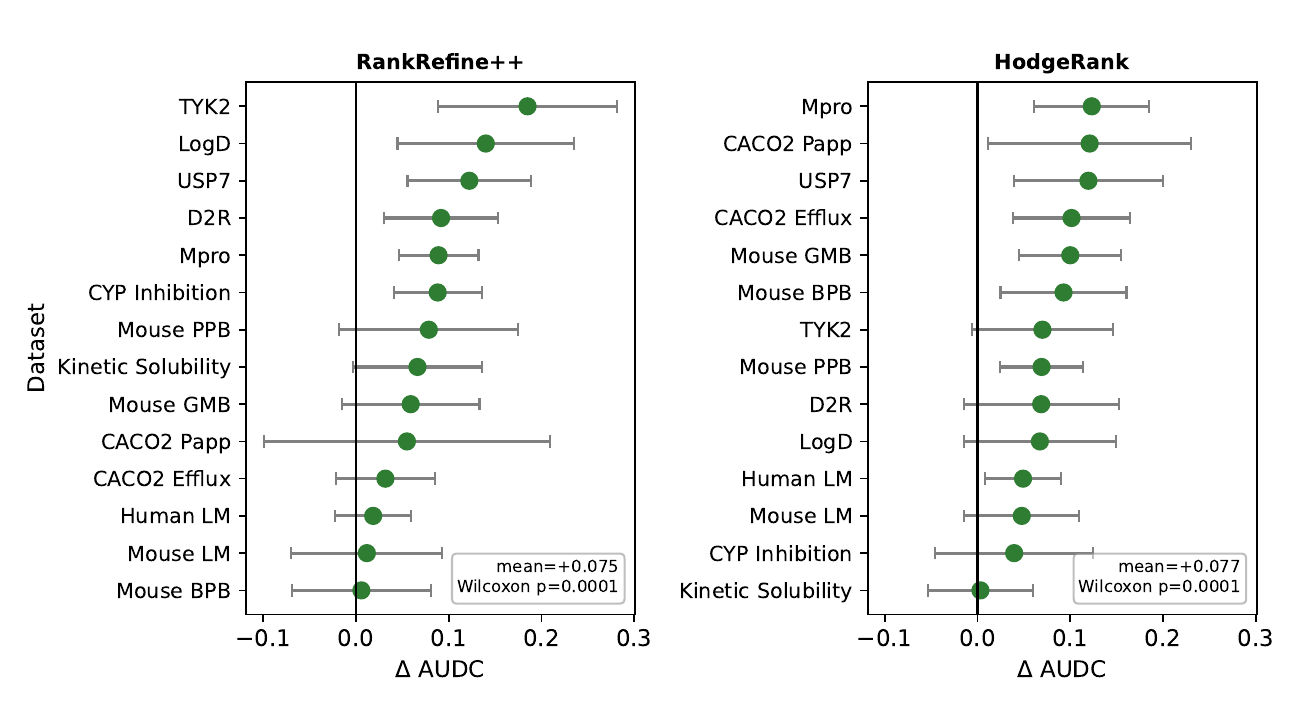}
    \caption{Paired difference in area under the top-10 discovery recall curve for RankRefine++ (left) and HodgeRank (right). Positive values indicate higher discovery performance for \methodname.
    }
    \label{fig:paired_dataset}
\end{figure}
Figure~\ref{fig:paired_dataset} shows the dataset-wise relative discovery performance of \methodname\ over RankRefine++ and HodgeRank.
Against RankRefine++ and HodgeRank, \methodname\ improves AUDC@10 on all drug discovery datasets, with statistically significant results under Wilcoxon signed-rank tests.
The results indicate that the aggregate improvements of \methodname\ in the main paper are broadly applicable across the drug discovery datasets.

\subsection{Engineering Discovery Campaign}
\begin{figure}[ht]
    \centering
    \setlength{\abovecaptionskip}{4pt}
    \includegraphics[width=0.95\linewidth]{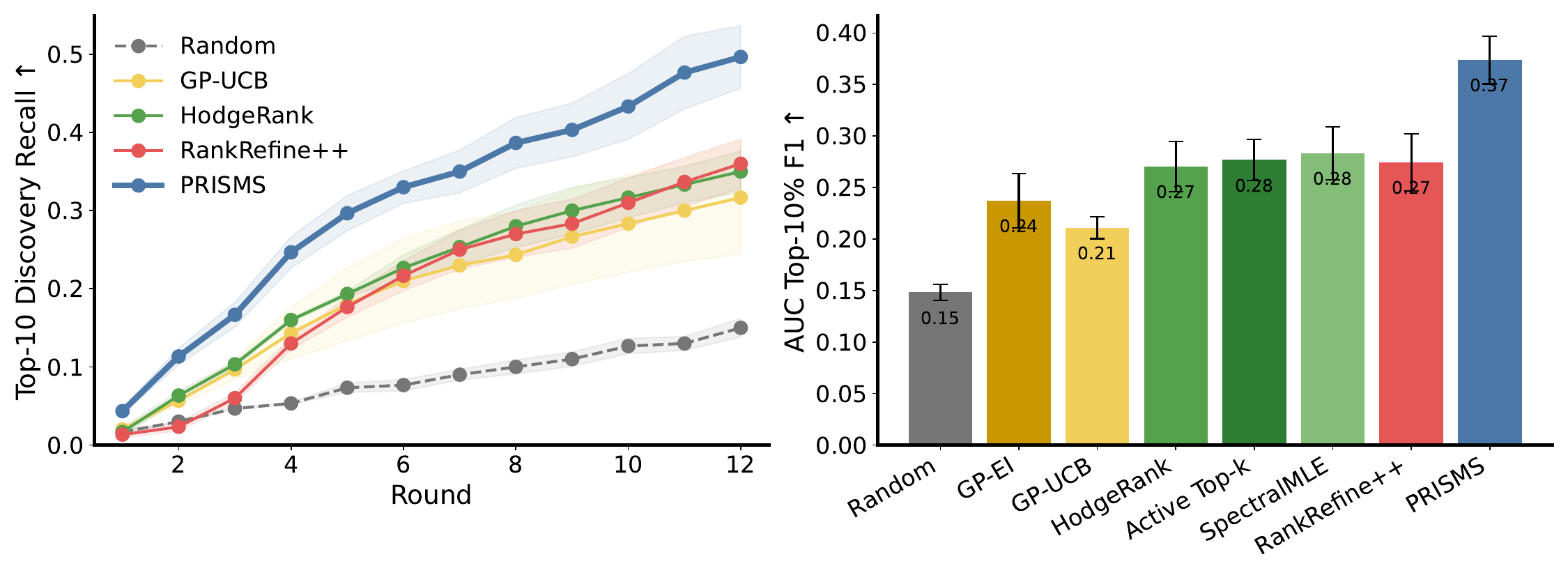}
    \caption{Aggregated results of 12-round scientific design discovery campaigns on three engineering datasets. We report the mean and standard error over 10 random seeds. \textbf{Left.} Top-10 discovery recall over 12 rounds. \textbf{Right.} AUC of Top-10\% F1 score.}
    \label{fig:engineering_screening}
\end{figure}
The detailed results for engineering discovery campaign are shown in Figure~\ref{fig:engineering_screening}.

\subsection{Ablation Study}
\begin{table}[t]
    \centering
    \caption{Ablation of PRISMS components and fidelity stages. We report the 24-round mean area under the discovery recall curve (AUDC) across 14 drug-discovery datasets, repeated for 10 random seeds, with standard error of the mean. Relative change is computed with respect to the default two-stage PRISMS configuration.
    }
    \label{tab:full_ablation}
    \small
    \begin{tabular}{lcc}
        \toprule
        Method 
        & AUDC@10 $\uparrow$ 
        & Relative change \\
        \midrule
        PRISMS 2 stage              & $0.527 \pm 0.049$ & --- \\
        \midrule
        No unlabeled-unlabeled edges  & $0.487 \pm 0.048$ & $-7.6\%$ \\
        No edge uncertainty            & $0.502 \pm 0.047$ & $-4.6\%$ \\
        No outcome uncertainty         & $0.514 \pm 0.048$ & $-2.3\%$ \\
        Random escalation              & $0.486 \pm 0.050$ & $-7.7\%$ \\
        No lateral edges + random escalation
                                        & $0.481 \pm 0.051$ & $-8.7\%$ \\
        Random Stage-1 initialization  & $0.519 \pm 0.047$ & $-1.4\%$ \\
        \midrule
        PRISMS - 1 stage              & $0.440 \pm 0.051$ & $-16.5\%$ \\
        PRISMS - 3 stage              & $0.586 \pm 0.045$ & $+11.4\%$ \\
        \bottomrule
    \end{tabular}
\end{table}
The detailed results for the ablation study are shown in Table~\ref{tab:full_ablation}.

\section{Implementation Details}
\label{app:implementation}

\subsection{Graph construction}
At the beginning of each round, each unmeasured design is connected to all measured designs and its $k^U$ nearest unmeasured neighbors according to the absolute difference in prior estimates.
The topology remains fixed throughout the round.
Comparisons are stored based on the edge and ranker identity.

\subsection{Optimization.}
Experimentally validated property values are held fixed, and the objective in Equation~\ref{eq:joint_estimation} is optimized over unmeasured values using L-BFGS-B with an analytic gradient.
The first optimization is initialized from the supplied prior estimates.
After higher-fidelity comparisons are added, optimization is initialized from the preceding $(\ell - 1)$ stage's solution.
The implementation defaults to at most 300 iterations, with function and gradient tolerances of $10^{-12}$ and $10^{-8}$, respectively.

\subsection{Sparse Hessian construction.}
After optimization, we assemble the analytic Hessian of the objective at the current estimates.
Each observed comparison from ranker $g_s$ contributes
\begin{equation}
w_{ij}^{(s)}\vv_{ij}\vv_{ij}^{\mathsf T},
\qquad
w_{ij}^{(s)}
=
\frac{\lambda_1^{(s)}}{\tau^2}
p_{ij}(\hat y^{(\ell)})
\left(1-p_{ij}(\hat y^{(\ell)})\right).
\end{equation}
For a comparison between two unmeasured designs, this adds $w_{ij}^{(s)}$ to both diagonal entries and $-w_{ij}^{(s)}$ to the corresponding off-diagonal entries.
Comparisons against experimentally validated designs contribute only to the diagonal of the unmeasured designs.
The prior contributes $2\lambda_2 I$.

The Hessian is stored in compressed sparse column format.
Contributions from multiple rankers to the same matrix entry are summed.
Its off-diagonal sparsity follows the comparison graph among unmeasured designs.

\subsection{Computing edge uncertainty.}
We evaluate $\vv_{ij}^{\mathsf T}(H^{(\ell)})^{-1}\vv_{ij}$ without explicitly forming the inverse.
A sparse LU factorization of $H^{(\ell)}$ is computed once before scoring the eligible comparisons.
For each pair, we reuse this factorization to solve
\begin{equation}
H^{(\ell)}\mathbf{u}_{ij}=\vv_{ij},
\end{equation}
and compute
\begin{equation}
U_{ij}^{\mathrm{edge}}
=
\vv_{ij}^{\mathsf T}\mathbf{u}_{ij}.
\end{equation}
This evaluates the quadratic form through direct linear solves without storing a dense covariance matrix.

\subsection{LLM-generated Rule-based Ranker}
For the cheap ranker ensemble, the rule-based ranker is generated using Claude Sonnet 4.6.
An example of the LLM-generated rule-based ranker is provided below.
\medskip

\begin{codebox}
\begin{lstlisting}
def predict_expansionrx_lm_clint(
    train_features: dict, 
    test_features: dict,
) -> int:
    """Predict liver microsomal intrinsic clearance (CLint) comparison."""
    
    if train_features is None or test_features is None:
        return _random_binary()

    def clint_score(f):
        score = 0.0
        score += f['aromatic_rings'] * 0.45
        score += f['logp'] * 0.35
        score += f['num_heteroatoms'] * 0.2
        score += f['rotatable_bonds'] * 0.12
        score -= f['fraction_csp3'] * 0.3
        return score

    if clint_score(train_features) > clint_score(test_features):
        return 1
    else:
        return 0
\end{lstlisting}
\end{codebox}

\subsection{Dataset}
\begin{table}[ht]
    \centering
    \caption{Details of datasets used in the experiments.}
    \begin{tabular}{lcl}
    \toprule
    Dataset Name & Dataset Size& Target Property \\
    \cmidrule(lr){1-3}
    Caco-2 efflux & $3777$  & Caco-2 efflux ratio \\
    Caco-2 Papp & $3773$ & Caco-2 permeability  \\
    Human LM & $4541$  & microsomal clearance\\
    Kinetic solubility & $7298$ & kinetic solubility \\
    LogD & $7309$ & LogD \\
    Mouse LM & $5692$ & microsomal clearance \\
    Mouse BPB & $1426$& brain fraction unbound \\
    Mouse GMB & $431$& mouse gastrocnemius muscle binding \\
    Mouse PPB & $1756$ & plasma fraction unbound \\
    Octant CYP3A4 & $1084$ & CYP3A4 inhibition potency  \\
    \cmidrule(lr){1-3}
    D2R & $2502$ & binding affinity ($\mathrm{p}K_i$)  \\
    Mpro & $665$  & binding affinity ($\mathrm{pIC}_{50}$) \\
    TYK2 & $9997$ & binding affinity ($\mathrm{p}K_i$) \\
    USP7 & $4535$ & binding affinity ($\mathrm{pIC}_{50}$)  \\
    \cmidrule(lr){1-3}
    3D printed structure & $600$ & mechanical toughness  \\
    Dye lasers & $3458$ & peak score  \\
    Redoxmer & $1408$ & absorption wavelength deviation  \\
    \bottomrule
    \end{tabular}
    \label{tab:dataset}
\end{table}

\end{document}